\documentclass[journal,twoside,web]{ieeecolor}
\usepackage{tmi}
\usepackage{cite}

\usepackage{amsfonts}
\usepackage{graphicx}
\usepackage[table,xcdraw]{xcolor}
\usepackage{colortbl}

\usepackage{comment}
\usepackage{amsmath,amssymb}
\usepackage{xcolor}
\usepackage{amsmath}
\usepackage{amssymb}
\usepackage{latexsym}
\usepackage{url}
\usepackage{cite}
\usepackage{relsize}
\usepackage{multirow}
\usepackage{tabularx}
\usepackage{array}
\usepackage{afterpage}
\usepackage{ifthen}
\usepackage{algpseudocode,algorithmicx}
\usepackage{flushend}
\usepackage{epstopdf}
\usepackage{stackengine}
\usepackage{gensymb}
\usepackage{epsfig}
\usepackage{mathrsfs}
\usepackage{tablefootnote}
\usepackage{stmaryrd}
\usepackage{soul}
\usepackage[english]{babel} 
\usepackage{blindtext}
\usepackage{enumitem}
\usepackage{subfig}

\makeatletter
\let\NAT@parse\undefined
\makeatother
\usepackage[bookmarks=false]{hyperref} 
\hypersetup{
    colorlinks=true,
    linkcolor=red,
    filecolor=magenta,      
    urlcolor=blue,
    pdfstartview={FitH},
    citecolor=blue
    }
    
\usepackage[ruled,vlined]{algorithm2e}
\usepackage{pifont}
\usepackage[capitalize]{cleveref}
\crefname{section}{Sec.}{Secs.}
\Crefname{section}{Section}{Sections}
\Crefname{table}{Table}{Tables}
\crefname{table}{Tab.}{Tabs.}

\usepackage{booktabs}
\usepackage{makecell}
\usepackage{hhline}
\usepackage{courier}
\usepackage{lipsum}

\usepackage{amsmath,amsfonts,bm} 
\usepackage{caption}

\usepackage{bm,cuted}
\usepackage{amsfonts}
\usepackage{graphicx}
\usepackage{tabularx} 
\usepackage{amssymb}
\usepackage[safe]{tipa}
\usepackage{tipa}
\usepackage{multirow}
\usepackage{xcolor}
\usepackage{hyperref}
\usepackage{dsfont}
\usepackage{upgreek}
\usepackage{url}
\newcommand{\etal}{\textit{et al}}

\usepackage{pifont}
\makeatletter
\newcommand\HUGE{\@setfontsize\Huge{20}{30}}
\makeatother

\def\eqref#1{equation~\ref{#1}}

\def\1{\bm{1}}

\DeclareMathAlphabet{\mathsfit}{\encodingdefault}{\sfdefault}{m}{sl}
\SetMathAlphabet{\mathsfit}{bold}{\encodingdefault}{\sfdefault}{bx}{n}

\definecolor{boxcolor}{HTML}{B9E035}

\usepackage{textcomp}
\def\BibTeX{{\rm B\kern-.05em{\sc i\kern-.025em b}\kern-.08em
    T\kern-.1667em\lower.7ex\hbox{E}\kern-.125emX}}
\begin{document}

\title{CathAction: A Benchmark for Endovascular Intervention Understanding}
\author{B. Huang, T. Vo, C. Kongtongvattana, G. Dagnino, D. Kundrat, W. Chi, M. Abdelaziz, T. Kwok,  T. Jianu, \\ T. Do, H. Le, M. Nguyen, H. Nguyen, E. Tjiputra, Q. Tran, J. Xie, Y. Meng, B. Bhattarai, Z. Tan, H. Liu, \\ H.S. Gan, W. Wang, X. Yang, Q. Wang, J. Su, K. Huang, A. Stefanidis, M. Guo, B. Du, R. Tao,\\ M. Vu, G. Zheng, Y. Zheng, F. Vasconcelos, D. Stoyanov, D. Elson, F.R y Baena, A. Nguyen\\ \vspace{1ex} \small \href{https://airvlab.github.io/cathaction/}{https://airvlab.github.io/cathaction/}
}



\twocolumn[{%
\renewcommand\twocolumn[1][]{#1}%
\maketitle

\begin{center}
  \centering
  \vspace{-14ex}
  \captionsetup{type=figure}
  \Large
\resizebox{\linewidth}{!}{
\setlength{\tabcolsep}{2pt}
\begin{tabular}{ccccc}
\shortstack{\includegraphics[width=0.99\linewidth]{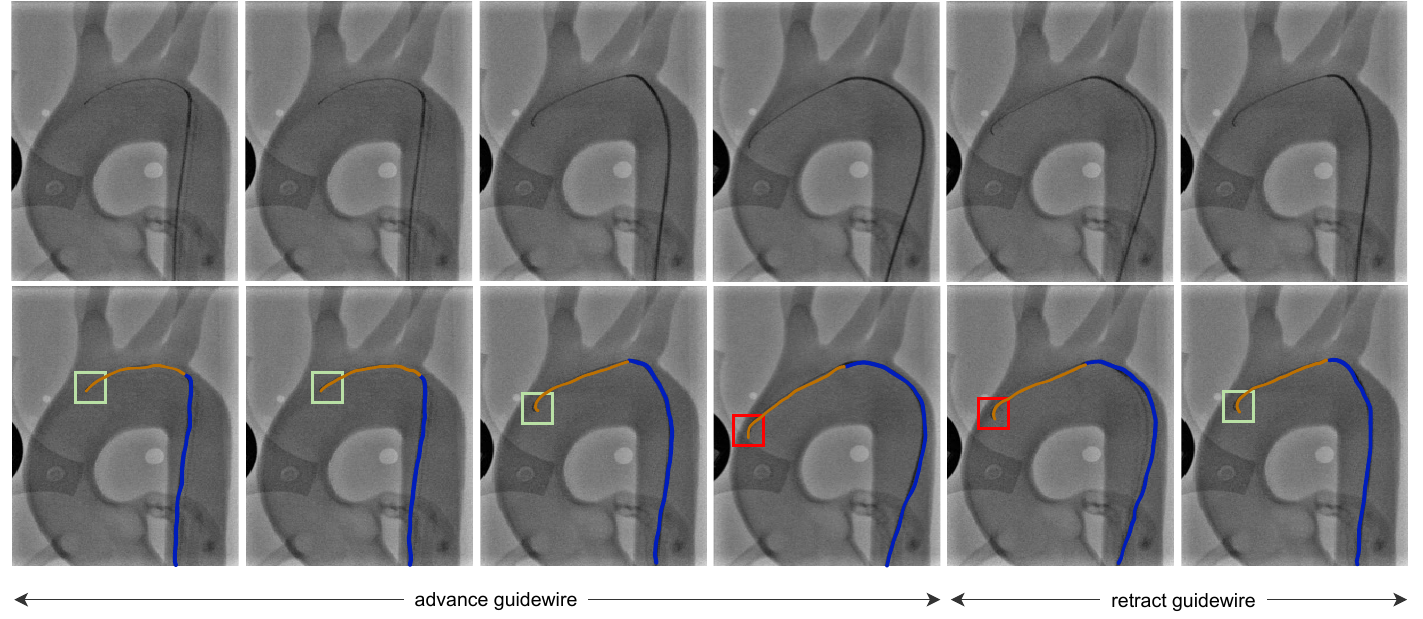}}\\[3pt]
\end{tabular}
}
\vspace{-1ex}
    \captionof{figure}{We present CathAction, a large-scale dataset for endovascular interventions. The top row shows an example X-ray sequence. The bottom row shows ground truth labels. The catheter is in \textcolor[HTML]{001DBC}{blue color}, the guidewire is in \textcolor[HTML]{BD7000}{dark yellow}, and the \textcolor[HTML]{FF0000}{red box} ${\color{red}\square}$ indicates the collision of the tip with the blood vessel, while the \textcolor[HTML]{B9E035}{green box} ${\color{boxcolor}\square}$ indicates no collision.}
    \label{fig:IntroVis}
\end{center}%
}]

{\renewcommand*{\thefootnote}{\fnsymbol{footnote}}\stepcounter{footnote}%
  \footnotetext{BH, TK, WC, MA, DK, DE, and FB are with Imperial College London; TV and HL are with FPT Software AI Center; MN and HN are with Vietnam National University; QT and ET are with AIOZ Ltd.; YM is with University of Exeter; BB is with University of Aberdeen; TR and GZ is with Shanghai Jiao Tong University; GD is with University of Twente; MG is with Wuhan University of Technology; B. Du is with Wuhan University, MV is with TU Wien; XY, QW, KH, WW, ZT, HSG, JS, AS, HL is with XJTLU, FV, DS are with UCL; TJ, TD, CK, JX, YZ, and AN are with University of Liverpool.\\
  The three first authors contribute equally.}}
\setcounter{footnote}{0}

\begin{abstract} 
Real-time visual feedback from catheterization analysis is crucial for enhancing surgical safety and efficiency during endovascular interventions. However, existing datasets are often limited to specific tasks, small scale, and lack the comprehensive annotations necessary for broader endovascular intervention understanding. To tackle these limitations, we introduce CathAction, a large-scale dataset for catheterization understanding. Our CathAction dataset encompasses approximately 500,000 annotated frames for catheterization action understanding and collision detection, and 25,000 ground truth masks for catheter and guidewire segmentation. For each task, we benchmark recent related works in the field. We further discuss the challenges of endovascular intentions compared to traditional computer vision tasks and point out open research questions. We hope that CathAction will facilitate the development of endovascular intervention understanding methods that can be applied to real-world applications. 
\end{abstract}
\vspace{-4ex}


\section{Introduction} \label{Sec:Intro}

\begin{table*}[t]

\centering
\resizebox{\textwidth}{!}{
\begin{tabular}{l@{\hskip 0.2in} l l r @{\hskip 0.1in} l l c l}
\toprule
\thead{\textbf{Dataset}} & \thead{\textbf{Collection}}  & \thead{\textbf{Type}}  & \thead{\textbf{\#Frames}} & \thead{\textbf{Source}} & \thead{\textbf{Annotation}} & \thead{\textbf{Public}} & \thead{\textbf{Task}} \\ 
\midrule
\midrule
Barbu \textit{et al.}~\cite{barbu2007hierarchical}  
& X-ray & Video & 535  & Real & Manual & No & Segmentation \\ \cmidrule[.001mm]{1-8}
Wu \textit{et al.}~\cite{wu2014fast}             
& 3D Echo & Video & 800  & Real & Manual & No & Segmentation\\  \cmidrule[.001mm]{1-8}
Ambrosini \textit{et al.}~\cite{ambrosini2017fully} 
& X-ray & Image  & 948  & Real & Manual & No & Segmentation \\  \cmidrule[.001mm]{1-8}
Mastmeyer \textit{et al.}~\cite{mastmeyer2017model}  
& 3D MRI & Image  & 101 & Real & Manual & No & Segmentation \\  \cmidrule[.001mm]{1-8}
Yi \textit{et al.}~\cite{yi2020automatic} 
& X-ray & Image  & 2,540 & Synthesis & Automatic & No & Segmentation\\  \cmidrule[.001mm]{1-8}
Nguyen \textit{et al.}~\cite{nguyen2020end} 
& X-ray & Image  & 25,271  & Phantom & Semi-Auto & No & Segmentation  \\  \cmidrule[.001mm]{1-8}
Danilov \textit{et al.}~\cite{danilov2023use} 
& 3D Ultrasound & Video  & 225 & Synthetic & Manual & No & Segmentation  \\  

\cmidrule{1-8}

Delmas \textit{et al.}~\cite{delmas2015three} 
& X-ray & Image  & 2,357 & Simulated & Automatic & No & Reconstruction\\  \cmidrule[.001mm]{1-8}
Brost \textit{et al.}~\cite{brost2010catheter} 
& X-ray & Image  & 938 & Clinical & Semi-Auto & No & Tracking\\  \cmidrule[.001mm]{1-8}
Ma \textit{et al.}~\cite{ma2010real} 
& X-ray, CT & Image & 1,048 & Clinical & Manual & No & Reconstruction\\  \cmidrule[.001mm]{1-8}
\cmidrule{1-8}
\textbf{CathAction} (ours) 
& X-ray & Video & 500,000+ & Phantom & Manual & Yes & \makecell[l]{Segmentation \\ Action Understanding \\ Collision Detection} \\ 
\bottomrule
\end{tabular}}
\vspace{1ex}
\caption{Endovascular intervention datasets comparison.}
\label{tab:DatasetsComparison}
\end{table*}

Cardiovascular diseases are one of the most common causes of death worldwide
~\cite{roth2018global}. Endovascular intervention is now
the gold standard of treatment for cardiovascular diseases~\cite{schneider2019endovascular} and is increasingly used due to its significant advantages compared to the traditional open surgery approach, such as smaller incisions, less trauma, reduced risks of developing comorbidities for patients~\cite{simaan2018medical}. Endovascular interventions involve maneuvering small and long medical instruments, i.e., \textit{catheter} and \textit{guidewire}, within the vasculature through small incisions to reach targeted areas for treatment delivery, such as artery stenting, tissue ablation, and drug delivery~\cite{rafii2014current}. However, such tasks have high technical requirements on surgeons' skills and the main challenge lies in avoiding collisions with the vessel wall, which can result in severe 
consequences, e.g., perforation, hemorrhage, and organ failure~\cite{molinero2019haptic}. In practice, surgeons can only rely on 2D X-ray images when performing the task in a 3D human body environment~\cite{abdelaziz2019toward}, which causes a significant challenge in safely controlling the catheter and guidewire.

Recently, learning methods for computer-assisted intervention systems have emerged for diverse tasks~\cite{huang2022self,qiu2022guidewire,huang2022simultaneous}. Numerous methodologies have tackled challenges in endovascular interventions, including catheter and guidewire segmentation~\cite{ambrosini2017fully,nguyen2020end}, vision-based force sensing~\cite{dagnino2018haptic}, learning from demonstration~\cite{chi2020collaborative}, and skill training assistance~\cite{jianu2022cathsim}. Additionally, various deep learning approaches have been proposed for specific tasks in endovascular interventions, such as instrument motion recognition in X-ray sequences~\cite{bostowards}, interventionalist hand motion recognition~\cite{ma2020irregular,hisey2020recognizing,akinyemi2023interventionalist,wang2023learning}, and collision detection~\cite{qiu2022guidewire,guo2021new,lyu2020cnn,zhao2020novel}. However, due to difficulties inherent in acquiring medical data, most of these methods rely on synthetic data~\cite{yi2020automatic} or small and private datasets~\cite{breininger2018multiple,breininger2018intraoperative}. Consequently, despite the critical nature of interventions, current methods have not fully capitalized on recent advancements in deep learning, which typically require large-scale training data.

Over the years, several datasets for endovascular intervention have been introduced~\cite{barbu2007hierarchical,wu2014fast,ambrosini2017fully,mastmeyer2017model,vlontzos2018deep,yi2020automatic,nguyen2020end,danilov2023use}. Table~\ref{tab:DatasetsComparison} shows a detailed comparison between current endovascular intervention datasets. However, present endovascular intervention datasets share common limitations. First, these datasets have relatively small sizes in terms of the number of images since collecting real-world medical data is an expensive procedure. Second, due to the privacy challenge in the medical domain, most of the current endovascular intervention datasets are kept private. Finally, existing datasets are created for one particular task such as segmentation, which do not support other important tasks in endovascular interventions such as collision detection or action understanding. 
To address these issues, we present \textbf{CathAction}, a large-scale dataset encompassing several endovascular intervention tasks such as segmentation, collision detection, and action understanding. To our knowledge, CathAction represents the largest and most realistic dataset specifically tailored for surgical equipment catheter and guidewire tasks.

In summary, we make the following contributions:
\begin{itemize}
   \item We introduce CathAction, a large-scale dataset for endovascular interventions. Our dataset provides manually labeled ground truth for segmentation, action understanding, and collision detection.
   \item We benchmark key tasks in endovascular interventions, including catheterization anticipation, recognition, segmentation, and collision detection.
   \item We discuss the challenges and open questions in endovascular intervention. Our code and dataset are publicly available. 
\end{itemize}

\section{Related Work}

\textbf{Endovascular Intervention Dataset.} 
Several endovascular intervention datasets have been introduced~\cite{barbu2007hierarchical,wu2014fast,ambrosini2017fully,mastmeyer2017model,vlontzos2018deep,yi2020automatic,nguyen2020end,danilov2023use}. Barbu~\etal.~\cite{barbu2007hierarchical} proposed a dataset that effectively localizes the entire guidewire and thoroughly validated it using a traditional threshold-based method. Other datasets, such as ~\cite{ambrosini2017fully,mastmeyer2017model,vlontzos2018deep,nguyen2020end} consider fluoroscopy videos at the image level, with mask annotations for each frame from the fluoroscopy videos. For instance, Ambrosini~\etal.~\cite{ambrosini2017fully} developed a dataset with $948$ annotated mask segmentation considering instances of both catheter and guidewire as one class. Similarly, Mastmeyer~\etal.~\cite{mastmeyer2017model} collected and annotated a dataset with $101$ segmentation masks for the real catheter from 3D MRI data. More recently, Nguyen~\etal.~\cite{nguyen2020end} proposed a dataset that considers both catheter and guidewire as one class. Overall, most of these datasets have limitations in terms of size, task categories, and focus. To overcome the limitations of existing endovascular intervention datasets, we introduce CathAction, a large-scale dataset with a variety of tasks, including catheter and guidewire segmentation, collision detection, and catheter action recognition and anticipation. The CathAction dataset enables the development of more accurate and reliable deep learning methods for endovascular interventions.

\textbf{Catheterization Action Understanding.} Deep learning techniques have demonstrated notable achievements in endovascular intervention action understanding~\cite{jianu2024autonomous}. Jochem~\etal~\cite{bostowards} presented one of the first works utilizing deep learning for catheter and guidewire activity recognition in fluoroscopy sequences. Subsequently, deep learning-based approaches~\cite{ma2020irregular,hisey2020recognizing,akinyemi2023interventionalist,wang2023learning} have gained prominence as the most widely utilized solution for interventionalist hand motion recognition. For instance, Akinyemi~\etal.~\cite{akinyemi2023interventionalist} introduced a deep learning model based on convolutional neural networks (CNNs) that incorporates convolutional layers for automatic feature extraction and identifies operators' actions. Additionally, Wang~\etal.~\cite{wang2023learning} proposed a multimodal fusion architecture for recognizing eight common operating behaviors of interventionists. Despite the extensive research on deep learning methods for endovascular intervention, it comes with the limitation of medical data: most of these methods use synthetic data~\cite{yi2020automatic} or small and private datasets~\cite{breininger2018multiple,breininger2018intraoperative}. This leads to the fact that although the intervention is a crucial procedure, it has not benefited from the recent development of deep learning methods, where large-scale training data are usually required~\cite{nguyen2020end}. 

\textbf{Catheter and Guidewire Segmentation}. Catheter and guidewire segmentation is crucial for real-time endovascular interventions~\cite{nguyen2020end}. Many methods have been proposed to address the challenges of catheter and guidewire segmentation~\cite{ambrosini2017fully,nguyen2020end,yi2020automatic,danilov2023use,yi2020automatic}. The outcomes of catheter and guidewire can be applied in vision-based force sensing~\cite{dagnino2018haptic}, learning from demonstration~\cite{chi2020collaborative}, or skill training assistance~\cite{jianu2022cathsim} applications. Traditional methods for catheterization segmentation adopt thresholding-based methods and do not generalize well on X-ray data~\cite{nguyen2020end}. Deep learning methods can learn meaningful features from input data, but it is challenging to apply to catheter segmentation due to the lack of real X-ray data and the tediousness of manual ground truth labeling~\cite{vlontzos2018deep}. Many current learning-based techniques for catheter segmentation and tracking are limited to training on small-scale datasets or synthetic data due to the challenges of large-scale data collection~\cite{yi2020automatic,vlontzos2018deep,mastmeyer2017model,ambrosini2017fully}. Our dataset provides manual ground truth labels for both the catheter and guidewire, hence offering a substantial development for catheter and guidewire segmentation.

\textbf{Collision Detection.} Collision detection is a crucial task in endovascular interventions to ensure patient safety~\cite{fischer2023sensorized,zhang2021magnetorheological}. Several attempts have been made to incorporate deep learning models into collision detection, but these methods have focused on identifying risky action in simulated datasets~\cite{qiu2022guidewire,guo2021new,lyu2020cnn,zhao2020novel}. While existing methods can be useful for identifying potential hazards, they cannot localize the position of collisions or provide visual feedback. Additionally, these methods have not been widely used in real-world settings due to the lack of annotated bounding boxes for collision of guidewire tips with vessel walls~\cite{qiu2022guidewire}. Our dataset addresses this limitation by providing annotated bounding boxes for collision events in both phantom and real-world data. This enables the development of deep learning models that can detect collisions in a real-time manner and provide visual or haptic feedback to surgeons~\cite{dagnino2018haptic}.

\section{The CathAction Dataset}
\label{Surgical_dataset}
This section introduces the CathAction dataset. Specifically, we describe the data collection process and annotation pipeline. We then present statistics regarding different aspects of our large-scale dataset.

\subsection{Data Collection}
\label{collect}
Given that endovascular intervention constitutes a medical procedure, the acquisition of extensive human data is often impractical and time-consuming due to privacy constraints. To address this challenge, we suggest an alternative approach involving the collection of data from two distinct sources: \textit{i)} utilizing vascular soft silicone phantoms modeled after the human body and \textit{ii)} employing animal subjects, specifically pigs. The selection of pigs as an alternative is justified by their vascular anatomy being widely acknowledged as highly analogous to that of humans~\cite{abdelaziz2019toward}.

\textbf{Ethic.} Since our data collection involves experiments with radiation sources (X-ray radiology fluoroscopic systems) and live animals, all relevant ethical approvals were obtained in advance the collection process. The human subjects who collect the data are well-trained and professional endovascular surgeons, wearing a protected suit as a daily practice in the hospital. 

\begin{figure}[!ht]
\centering
\subfloat[Silicon phantom]{\label{fig:convex_hull_construction}\includegraphics[height=0.45\linewidth]{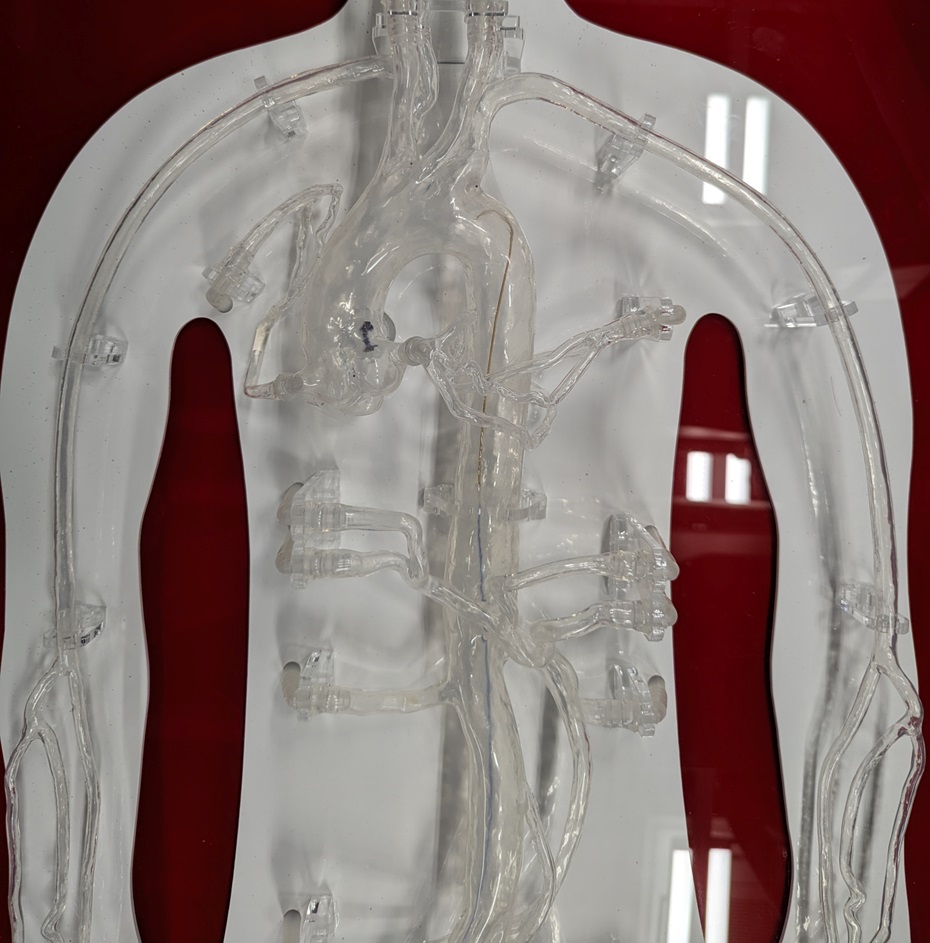}}
\hspace{3ex}
\subfloat[Data collection setup]{\label{fig:grasp_pose_evaluation}\includegraphics[height=0.45\linewidth]{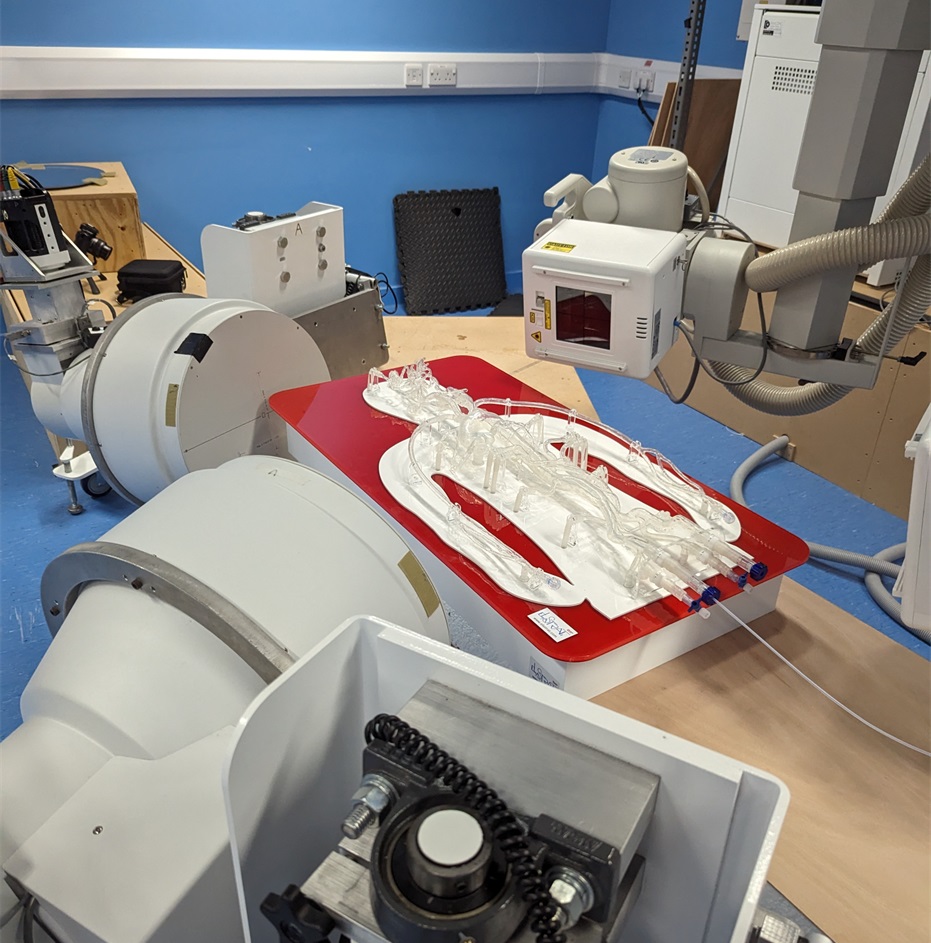}}
\caption{The human silicon phantom model (a), and data collection setup in the operating room (b).}
\label{fig_data_collection_setup}
\end{figure}

\textbf{Phantom Setup.} To ensure that the data are collected from various models, we use five adult human aortic arch phantoms made of soft silicone, manufactured by Elastrat Ltd., Switzerland. To enhance the realism of the interaction between surgical tools and tissues, the phantoms are connected to a pulsatile pump to simulate the flow of normal human blood. All phantoms are placed beneath an X-ray imaging system to mimic a patient lying on an angiography table, preparing for an endovascular procedure.

\textbf{Animal Setup.} We use five live pigs as the subjects for data collection. The animal setup is identical to that of on human procedure. During the endovascular intervention, professional surgeons use an iodine-based contrast agent to enhance the visibility of specific structures or fluids within the body~\cite{lusic2103}. Iodine contrast agents are radiopaque, meaning they absorb X-rays, resulting in improved visibility of blood vessels, organs, and other structures such as the catheter and guidewire during the imaging process.

\begin{figure}[h]
    \centering
    \includegraphics[width=.99\linewidth, height=0.75\linewidth]{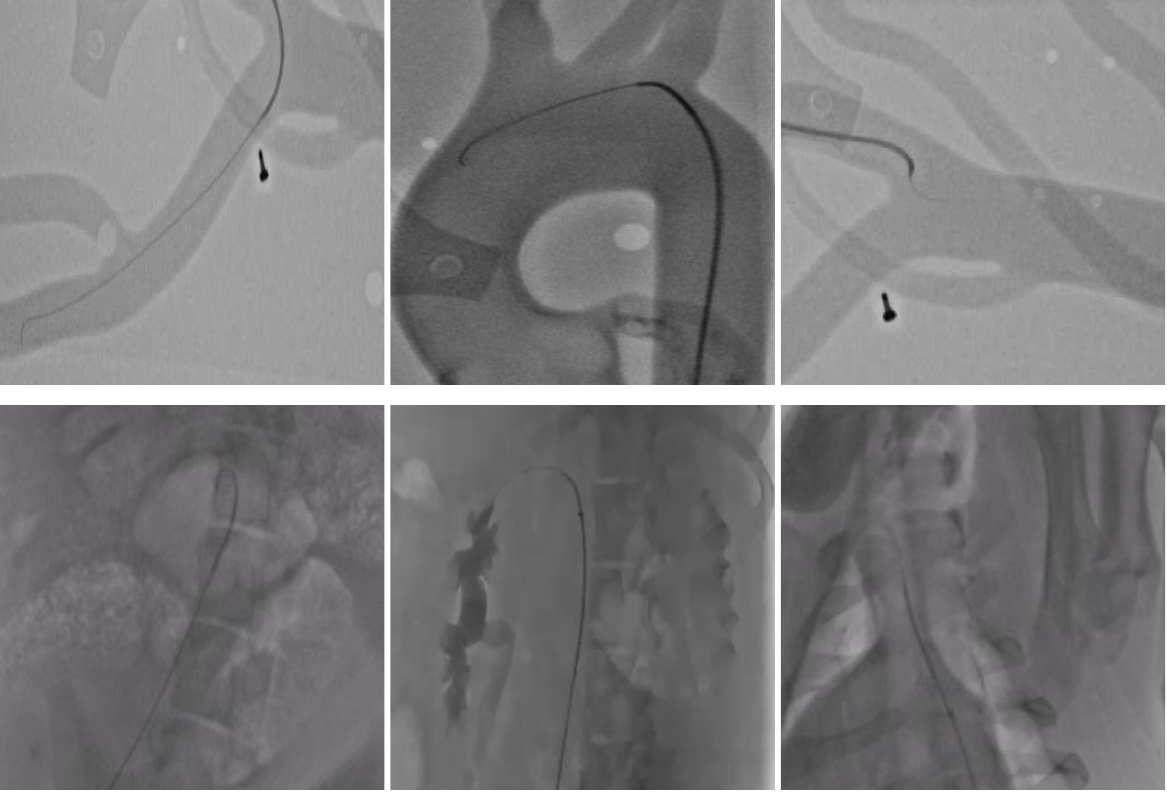}
    \caption{Example data collected with phantom models (top row) and animals (bottom row). Animal data are more challenging with less visible catheters or guidewires.}
    \label{fig_collected_data}
\end{figure}

\textbf{Data Collection.} Ten skilled professional surgeons are tasked with cannulating three arteries, namely the left subclavian (LSA), left common carotid (LCCA), and right common carotid (RCCA), using the commercial catheter and guidewire. Throughout each catheterization process, the surgeon operator activated the X-ray fluoroscopy using a pedal in the operating room. We develop a real-time image grabber to transmit the video feed of the surgical scene to a workstation. The experiments are conducted under two interventional radiology fluoroscopic systems, namely Innova 4100 IQ GE Healthcare and 
EMD Technologies Epsilon X-ray Generator. Fig.~\ref{fig_data_collection_setup} shows the data collection setup with human silicon phantoms and Fig.~\ref{fig_collected_data} visualizes the collected data with phantom models and real animals. From Fig.~\ref{fig_collected_data}, we can see that there is a huge \textit{domain gap} between data collected using phantom models and live animals.

\subsection{Data Annotation}
\textbf{Actions.} Based on the advice from expert endovascular surgeons, we define five classes to annotate catheterization actions. These classes belong to three groups: catheter (\texttt{advance catheter} and \texttt{retract catheter}), guidewire (\texttt{advance guidewire} and \texttt{retract guidewire}), and one action involving both the catheter and guidewire (\texttt{rotate}). We note that, in practice, the endovascular surgeons usually rotate both the catheter and guidewire simultaneously, hence we only use one rotation class. We utilize a free and open-source video editor~\cite{subtitleedit} to annotate the start and end times of each narrated action. All the fluoroscopy videos are processed at $500 \times 500$ resolution and $24$ frames per second (FPS). To ensure the quality of the annotation, all the groundtruth actions are manually checked and modified by an experienced endovascular surgeon. 

\textbf{Collision Annotation.} In practice, the collision between the catheter (or guidewire) with the blood vessel wall mostly happens via the tip of the instrument. Therefore, for each frame of the fluoroscopy video, we annotate the tip of the catheter (or guidewire) with a bounding box. There are two classes for the bounding boxes: \texttt{collision} (when the instrument collides with the blood vessel) or \texttt{normal} (when there is no collision). We exploited an open-source labeling and annotation tool~\cite{darklabel} to annotate the bounding boxes in each video. We note that the bounding box annotation can be done independently with the catheterization actions annotation, however, all the videos are encoded with $24$ FPS to ensure coherence in the dataset.

\textbf{Segmentation.} The combination of guidewire and catheter is common in endovascular interventions, where precise navigation through blood vessels is essential for the success of the procedure. Unlike most of the previous datasets that consider both catheter and guidewire as one class~\cite{nguyen2020end, vlontzos2018deep}, we manually label the catheter and the guidewire class separately in our dataset. Therefore, our segmentation groundtruth provides a more detailed understanding of endovascular interventions.

\subsection{Dataset Statistic}

\textbf{Overview.} As summarized in Table~\ref{tab:DatasetsComparison}, our CathAction is a large-scale benchmark for endovascular interventions. Our dataset consists of approximately $500,000$ annotated frames for action understanding and collision detection, and around $25,000$ ground-truth masks for catheter and guidewire segmentation. There are a total of $569$ videos in our dataset. Some of the collected video samples are illustrated in Fig.~\ref{fig_collected_data}. With these statistics, we believe CathAction is currently the largest, most challenging, and most comprehensive dataset of endovascular interventions.

\begin{figure}[h]
    \centering
    \includegraphics[width=0.95\linewidth]{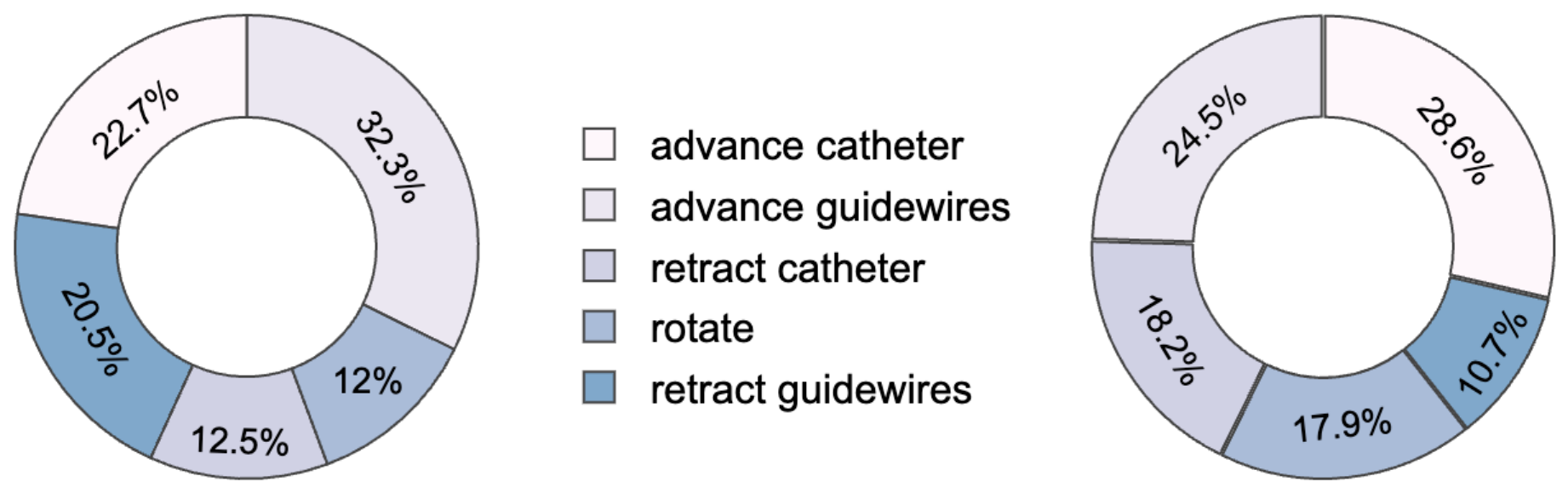}
    \caption{Distribution of the number of action classes in the CathAction dataset. Left-side: Distribution on real animal data. Right-side: Distribution on phantom data.}
    \label{statitics:static_1}
\end{figure}

\begin{figure}[h]
    \centering
    \includegraphics[width=0.99\linewidth]{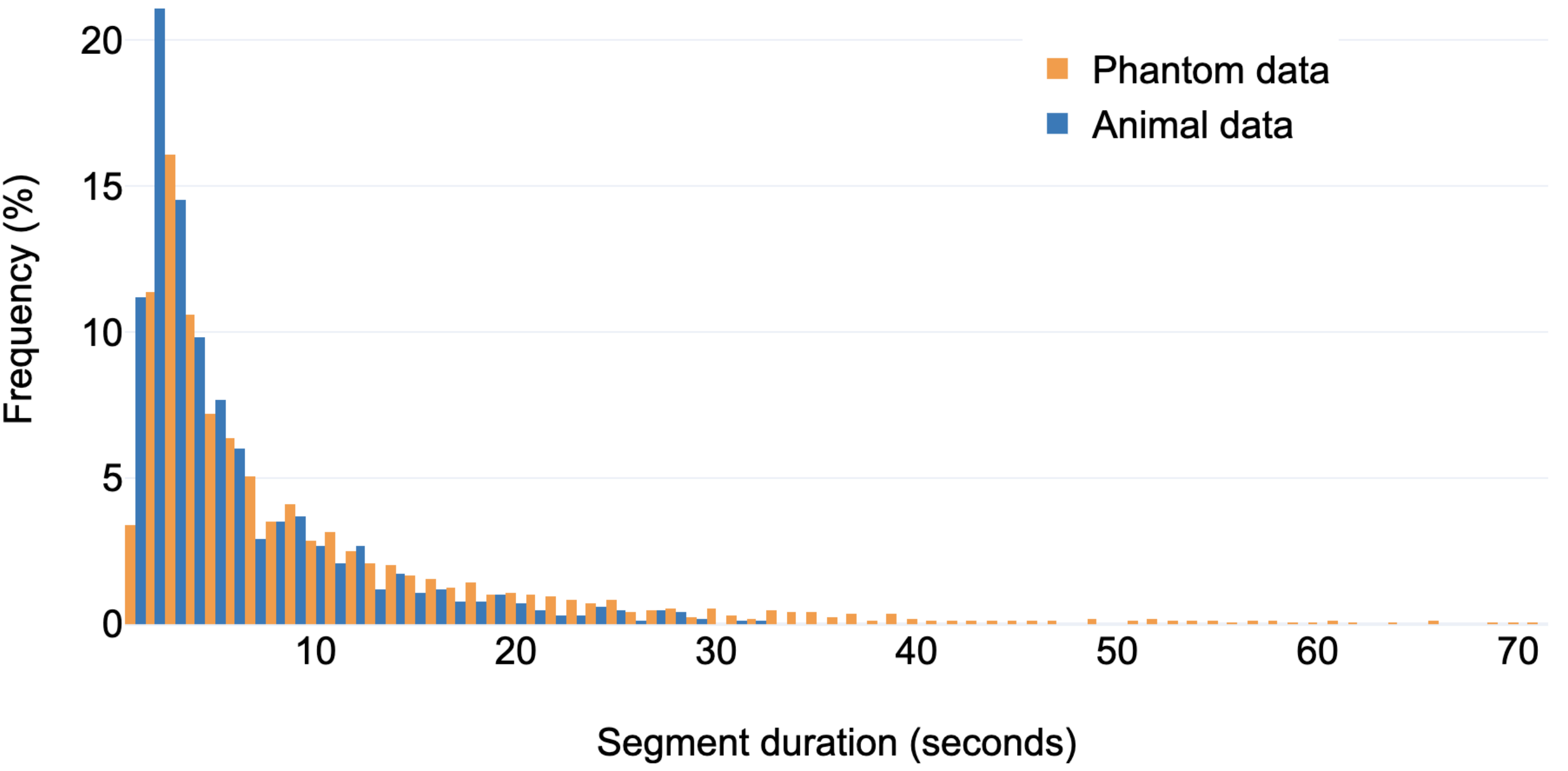}
    \caption{Duration distribution of segments' actions in the CathAction dataset, on real animal data and phantom data.}
    \label{statitics:static_2}
\end{figure}

\begin{figure}[h]
    \centering
    \includegraphics[width=0.98\linewidth]{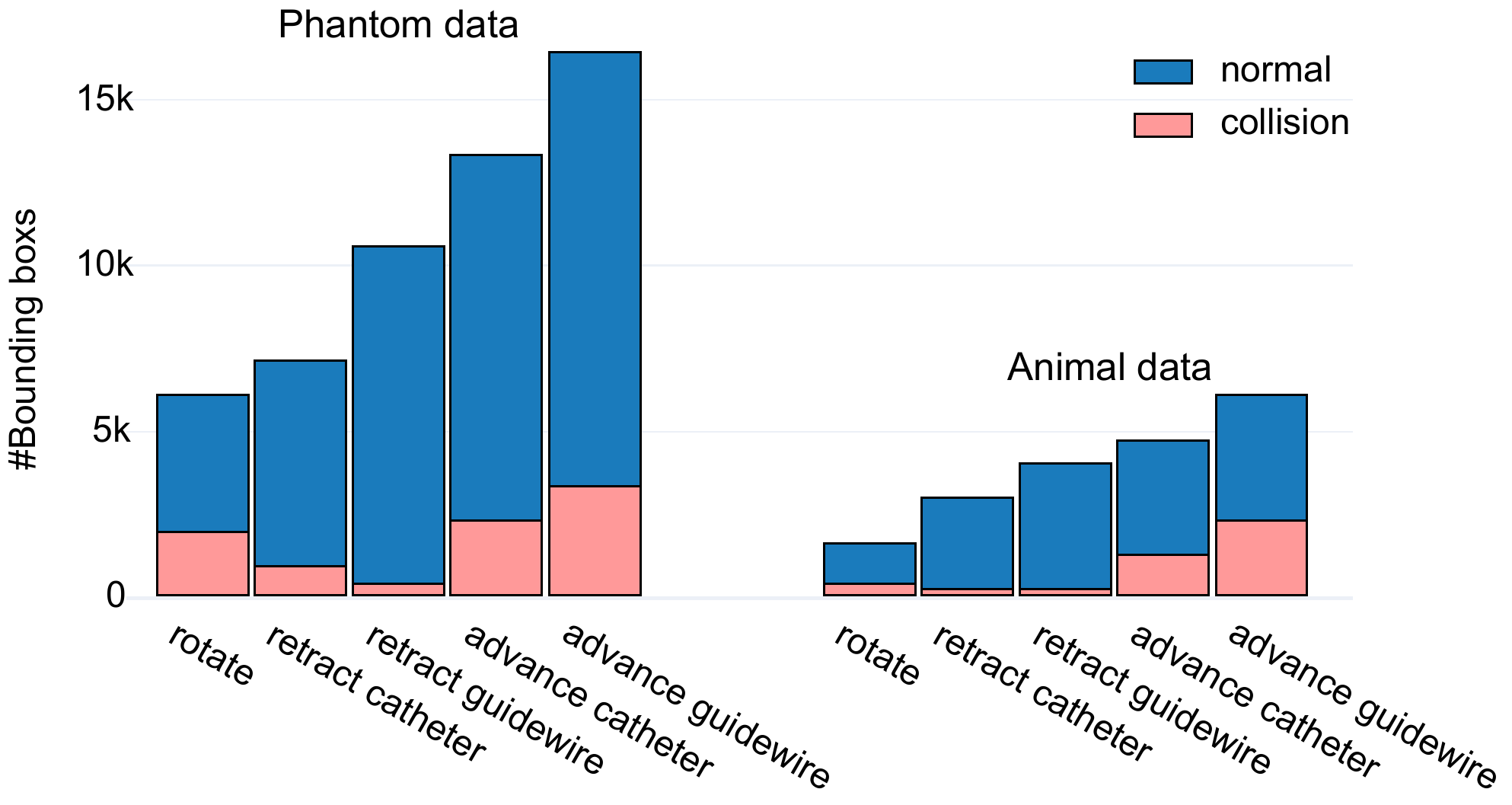}
    \caption{Comparison of the number of bounding box objects in real animal data and phantom data.}
    \label{statitics:static_box}
\end{figure}

\textbf{Statistics.} We annotate the CathAction dataset with a primary focus on catheters and guidewires. 
Fig.~\ref{statitics:static_1} provides an overview of the distribution of action classes in both animal and phantom data. Fig.~\ref{statitics:static_2} portrays the distribution of action segment lengths, illustrating the substantial variability in segment duration. Additionally, Fig.~\ref{statitics:static_box} visually compares the number of bounding boxes between phantom data and animal data. Notably, this comparison reveals a significant disparity between the counts of normal bounding boxes and collision boxes, an expected observation due to the infrequency of collision events in real-world scenarios.

\textbf{Adaptation Property.} Since we collect the data from two sources: phantom and real animals, there is a domain gap between the two types of data. Fig.~\ref{fig_collected_data} and Fig.~\ref{statitics:static_box} also show the adaptation property shared between the phantom data and the animal data. This distinctive domain gap feature renders the CathAction dataset a formidable benchmark for evaluating domain adaptation, which is a popular problem in the medical domain as collecting real human data is often infeasible~\cite{li2020domain,tobore2019deep,cios2002uniqueness,schlegl2014unsupervised}. Utilizing our CathAction dataset, we can develop domain adaptation techniques learning from synthetic or phantom data and effectively applying that knowledge to genuine animal/human data, effectively bridging the gap between controlled simulation data and real-world scenarios.

\section{Tasks and Benchmarks}
\label{sec_task}
In this section, we benchmark five tasks, including anticipation, recognition, segmentation, collision detection, and domain adaptation, to demonstrate the usefulness of the CathAction dataset. We then discuss the challenges and opportunities for improvement in each task.

\subsection{Catheterization Anticipation}
Given a sequence of frames, the goal of the anticipation task is to predict the action class for the next catheterization action. Following traditional anticipation tasks in computer vision~\cite{damen2018scaling}, we adopt two timing parameters: anticipation time ($\tau_a$) and observation time ($\tau_o$). Anticipation time indicates the amount of time we need to recognize an action, and observation time is the length of the video footage we need to look at before predicting an action. We predict the action class $c_a$ for the frames in the anticipation time $\tau_a$, given the frames in observation time $\tau_o$.

\textbf{Network and Training.} We utilize state-of-the-art action anticipation methods as baselines: CNN\&RNN~\cite{abu2018will}, RU-LSTM~\cite{abu2018will}, TempAggRe-Fusion~\cite{sener2020temporal}, AFFT~\cite{zhong2023anticipative}, and Trans-SVNet~\cite{jin2022trans}. The next action prediction is supervised using cross-entropy loss with labeled future actions. Similarly to~\cite{abu2018will,sener2020temporal,zhong2023anticipative}, we set $\tau_a=1s$, $\tau_o=1s$. We used a single Nvidia A100 GPU to train the models with a batch size of $64$. The training process lasted for $80$ epochs, and we started with an initial learning rate of $0.001$. We reduced the learning rate by a factor of $10$ after $30$ to 60 epochs. All other parameters are re-used from the baseline methods. We split roughly 80\% of the dataset for training and 20\% for testing. We utilize metrics presented in~\cite{zhao2017slac} to evaluate the results, including top-1 accuracy, precision, and recall. 

\begin{table}[h]
\centering
\renewcommand
\tabcolsep{4.5pt}
\hspace{1ex}
\vskip 0.1 in
\resizebox{\linewidth}{!}{
\begin{tabular}{@{}lcccc@{}}
\toprule
Baseline & Venues & Accuracy & Precision &Recall\cr 
\midrule
CNN~\cite{abu2018will} &CVPR 2018 & 28.98  & 30.14 &  29.76 \\ 
RNN~\cite{abu2018will}&CVPR 2018& 29.64  & 30.38  &30.44 \\
RU-LSTM~\cite{furnari2019would}& CVPR 2019& 35.08  & 34.29  & 34.77 \\
TempAggRe~\cite{sener2020temporal}& ECCV 2020 & 34.64  & 35.56  &34.71 \\ 
Trans-SVNet~\cite{jin2022trans} & IJCARS 2022 & 29.06& 19.67  & 20.28 \\
AFFT~\cite{zhong2023anticipative} & WACV 2023 & \textbf{37.91} & \textbf{36.87}  & \textbf{37.63} \\
\bottomrule
\end{tabular}
}
\caption{\label{tab:act_ant} Catheterization anticipation results on the CathAction dataset. All values are reported in percentages (\%).}

\end{table}

\begin{figure}[t]
    \centering
    \includegraphics[width=.99\linewidth]{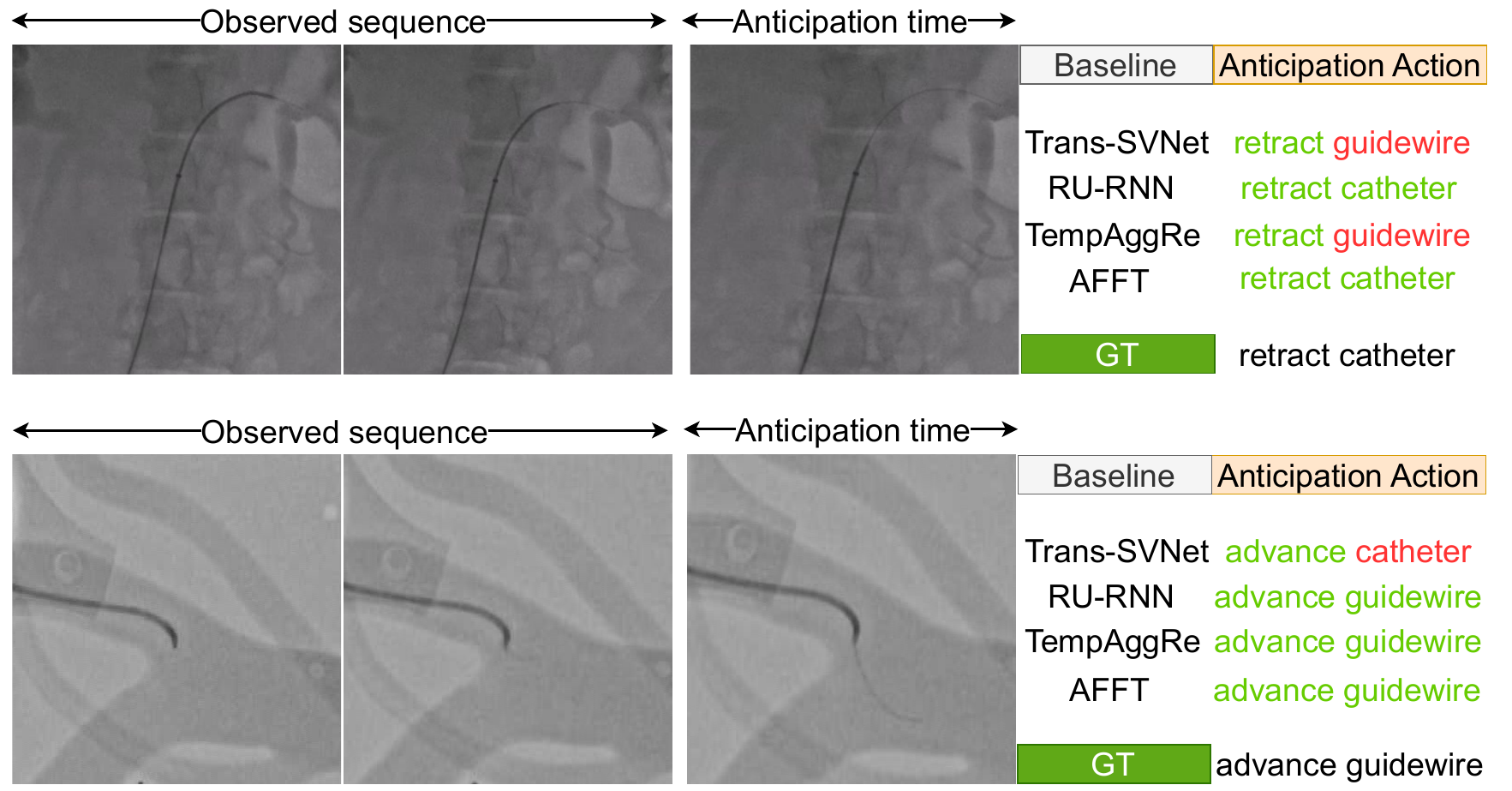}
    \caption{Qualitative catheterization prediction results. The predicted and ground truth of the next action are displayed on the right of each sample. The green color shows the correct prediction, and the red color shows the incorrect prediction.}
    \label{fig:ant_qualitative}
\end{figure}

\textbf{Results}. Table~\ref{tab:act_ant} shows the catheterization anticipation results of different baselines. This table shows that transformer-based methods show superior performance advantages over CNN or LSTM-based models. 
Qualitative results are illustrated in Fig.~\ref{fig:ant_qualitative}. We can see that transformer-based models can make more accurate predictions in challenging scenarios, especially when the catheter is moving quickly or when the occlusion is partially obscured. 

\textbf{Discussion.} Existing methods for catheterization anticipation still fall short of achieving high accuracy, presenting exciting opportunities for future research. The motion of the catheter and guidewire is fast, while their small and long thin body appearance poses a significant challenge on this task. Furthermore, real-time performance is crucial in this domain, as the surgeons need to receive real-time feedback from visual perception during the surgery.


\subsection{Catheterization Recognition}

Following the traditional action recognition task in computer vision~\cite{damen2018scaling}, in catheterization recognition, given an input video segment, our goal is to predict the action class for that video segment.

\textbf{Network and Training.} We explore state-of-the-art methods in action recognition to benchmark the catheterization recognition task, including TDN~\cite{wang2021tdn}, Video Swin Transformer~\cite{xia2022vision}, and BERT Pretraining of Video Transformers (BEVT)~\cite{wang2022bevt}. We train each model with two Nvidia A100 GPUs for $80$ epochs, with a mini-batch size of $512$. The initial learning rates were $0.01$ for the spatial stream and $0.001$ for the temporal stream, reduced by a factor of 10 at the $20$ and $40$ epochs. All
other parameters are re-used from the baseline methods.


\textbf{Results.} 
Table~\ref{tab:Recog} show the catheterization recognition results of three baseline methods: TDN~\cite{wang2021tdn}, Video Swin-Transformer~\cite{xia2022vision}, and BEVT~\cite{wang2022bevt}, on the CathAction dataset. TDN~\cite{wang2021tdn} with ResNet101 achieves the best top-1 accuracy of $62.5\%$ on five classes. We note that action recognition in endovascular intervention remains a challenging problem because the appearance of catheters and guidewires is relatively similar across different environments, while actions depend on the visual characteristics of the catheters and guidewires. 

\textbf{Discussion.} Compared to the anticipation task (Table~\ref{tab:act_ant}), catheterization recognition methods (Table~\ref{tab:Recog}) show higher accuracy. However, the overall performance is not yet significant enough to be applied to real-world applications. Further research can utilize more advanced techniques such as multi-modality learning~~\cite{tufek2019human,chen2017survey,ahmad2021inertial}, combining pre-operative data or synthesis data with transfer learning~\cite{raab2023modi,atasever2023comprehensive,luo2023source,himeur2023video,reddy2023synthetic} to improve the results. Furthermore, exploring the capabilities of large-scale medical foundation models is also an interesting research direction~\cite{moor2023foundation,qiu2023large,ma2023segment}.


\begin{table}[t]
\centering
\renewcommand
\tabcolsep{4.5pt}
\hspace{1ex}
\vskip 0.1 in
\resizebox{\linewidth}{!}{
\begin{tabular}{@{}lcccc@{}}
\toprule
Baseline & Venues & Accuracy & Precision &Recall\cr 
\midrule
TDN-ResNet50~\cite{wang2021tdn}  &CVPR 2021& 58.34&59.12&57.22    \\
TDN-ResNet101~\cite{wang2021tdn} &CVPR 2021& \textbf{62.50}&\textbf{61.89}& \textbf{62.77}  \\
Video Swin Transformer~\cite{xia2022vision} &CVPR 2022& 51.67 &52.14&51.24    \\
BEVT ~\cite{wang2022bevt} &CVPR 2022& 49.28&50.27&49.92 \\
\bottomrule
\end{tabular}
}
\caption{\label{tab:Recog} Catheterization recognition results on the CathAction dataset. All values are reported in percentages (\%).}

\end{table}


\subsection{Catheter and Guidewire Segmentation}
Catheter and guidewire segmentation is a well-known task in endovascular interventions~\cite{nguyen2020end}. In this task, we aim to segment the catheter and guidewire from the background. Unlike catheterization recognition or anticipation problem which takes a video as the input, in this segmentation task, we only use the X-ray image as the input. 

\begin{table}[h]
\centering
\renewcommand
\tabcolsep{4.5pt}
\hspace{1ex}
\vskip 0.1 in
\resizebox{\linewidth}{!}{
\begin{tabular}{@{}lcccccc@{}}
\toprule
Baseline & Dice Score & Jaccard Index & mIoU & Accuracy\cr 
\midrule
UNet~\cite{ronneberger2015u}       & 51.69    & 57.51     &  31.17    & 63.26\\ 
TransUNet~\cite{chen2021transunet} & 56.52    & 55.93     & 34.13    & 55.61 \\ 
SwinUNet~\cite{ambrosini2017fully} & 61.26    &  \textbf{59.54}     & 39.53    &  \textbf{76.60}\\
SSL~\cite{kongtongvattana2023shape} & 56.95    & 56.87     & 40.87    & 72.24\\
SegViT~\cite{zhang2022segvit}      & \textbf{63.47}    &{54.12}     & \textbf{42.48}    & {68.73} \\
\bottomrule
\end{tabular}}
\caption{\label{table:seg_results} Segmentation results on the CathAction dataset.}
\end{table}

\textbf{Network and Training.} We benchmark U-Net~\cite{ronneberger2015u}, Trans-UNet~\cite{chen2021transunet}, SwinUNet~\cite{ambrosini2017fully}, and SegViT~\cite{zhang2022segvit}. We follow the default training and testing configurations provided in the published papers. Following~\cite{ronneberger2015u,chen2021transunet,ambrosini2017fully,zhang2022segvit}, we use the Dice Score, Jaccard Index, mIoU, and Accuracy as the evaluation metrics in the segmentation task.

\textbf{Results.} Table~\ref{table:seg_results} shows the catheter and guidewire segmentation results. This table shows that the transformer-based networks such as TransUNet~\cite{chen2021transunet} or SegViT~\cite{zhang2022segvit} achieve higher accuracy than the traditional UNet~\cite{ronneberger2015u}. The SegViT~\cite{zhang2022segvit} that utilizes the vision transformer backbone shows the best performance, however, the increase compared with other methods is not a large margin.

\textbf{Discussion.} In contrast to traditional segmentation tasks in computer vision, which typically involve objects occupying substantial portions of an image~\cite{lin2014_COCO}, the segmentation of catheters and guidewires presents a considerably greater challenge. These elongated instruments possess extremely slender bodies, making their spatial presence in the image less pronounced. Moreover, the unique characteristics of X-ray images contribute to the potential misidentification of catheters or guidewires as blood vessels. Addressing these challenges in future research endeavors is imperative to enhance the accuracy of segmentation outcomes.

\subsection{Collision Detection}
Detecting the collision of the tip of the catheter or guidewire to the blood vessel wall is an important task in endovascular intervention~\cite{fischer2023sensorized,zhang2021magnetorheological}. We define the collision detection task as an object detection problem. In particular, the tip of the catheter or guidewire of all frames in our dataset is annotated with a bounding box. Each bounding box shares the class of ether \texttt{colision} when the tip collides with the blood vessel, or \texttt{normal} when there is no collision with the blood vessel.

\textbf{Network and Training.} We use YOWO~\cite{kopuklu2019you}, YOWO-Plus~\cite{yang2022yowo}, STEP~\cite{yang2019step}, and HIT~\cite{faure2023holistic}. Since the bounding boxes in our ground truth have relatively small sizes, we also explore tiny object detection methods such as Yolov~\cite{shi2023yolov} and EFF~\cite{gong2021effective}. The training process starts with a learning rate of $0.0003$, which is then decreased by a factor of $10$ after $20$ epochs, concluding at $80$ epochs. We train all methods with a mini-batch size of $4$ on an Nvidia A100 GPU. The average precision (AP) and mean average precision (mAP) are used to evaluate the detection results.

\begin{table}[h]
\centering
\renewcommand
\tabcolsep{4.5pt}
\hspace{1ex}
\vskip 0.1 in
\resizebox{\linewidth}{!}{
\begin{tabular}{@{}lcccccc@{}}
\toprule
  & \multicolumn{3}{c}{\textbf{AP}} & \multicolumn{3}{c}{\textbf{mAP}} \\
\cmidrule(lr){2-4}\cmidrule(lr){5-7}
Baseline &  
Collision & Normal  & Mean & Collision & Normal & Mean \cr 
\midrule
STEP~\cite{yang2019step}& 7.79 & 11.21 & 10.98  & 6.92 & 11.29 & 9.08  \\
YOWO~\cite{kopuklu2019you}& 8.32 & 12.18 & 11.73  & 7.46 & 12.28 & 9.92  \\
YOWO-Plus~\cite{yang2022yowo}& 8.92 & 12.23 & 11.77  & 7.86 & 12.48 & 10.28  \\
HIT~\cite{faure2023holistic}& 9.37  & 12.74& 12.14  & 8.18 & 12.72 & 10.81  \\
\hline
Yolov$^*$~\cite{shi2023yolov} &12.30 &21.08&15.89&11.88&20.04&14.11 \\
EFF$^*$~\cite{gong2021effective} &\textbf{13.70} &\textbf{22.10}&\textbf{16.91}&\textbf{12.14}&\textbf{20.78}&\textbf{14.88} \\
\bottomrule
\end{tabular}}
\caption{\label{table:collision_results} Collision detection results on the CathAction dataset. The symbol ($^*$) denotes tiny object detectors.}
\end{table}

\begin{figure}[t]
    \centering
    \includegraphics[width=.99\linewidth, height=0.55\linewidth]{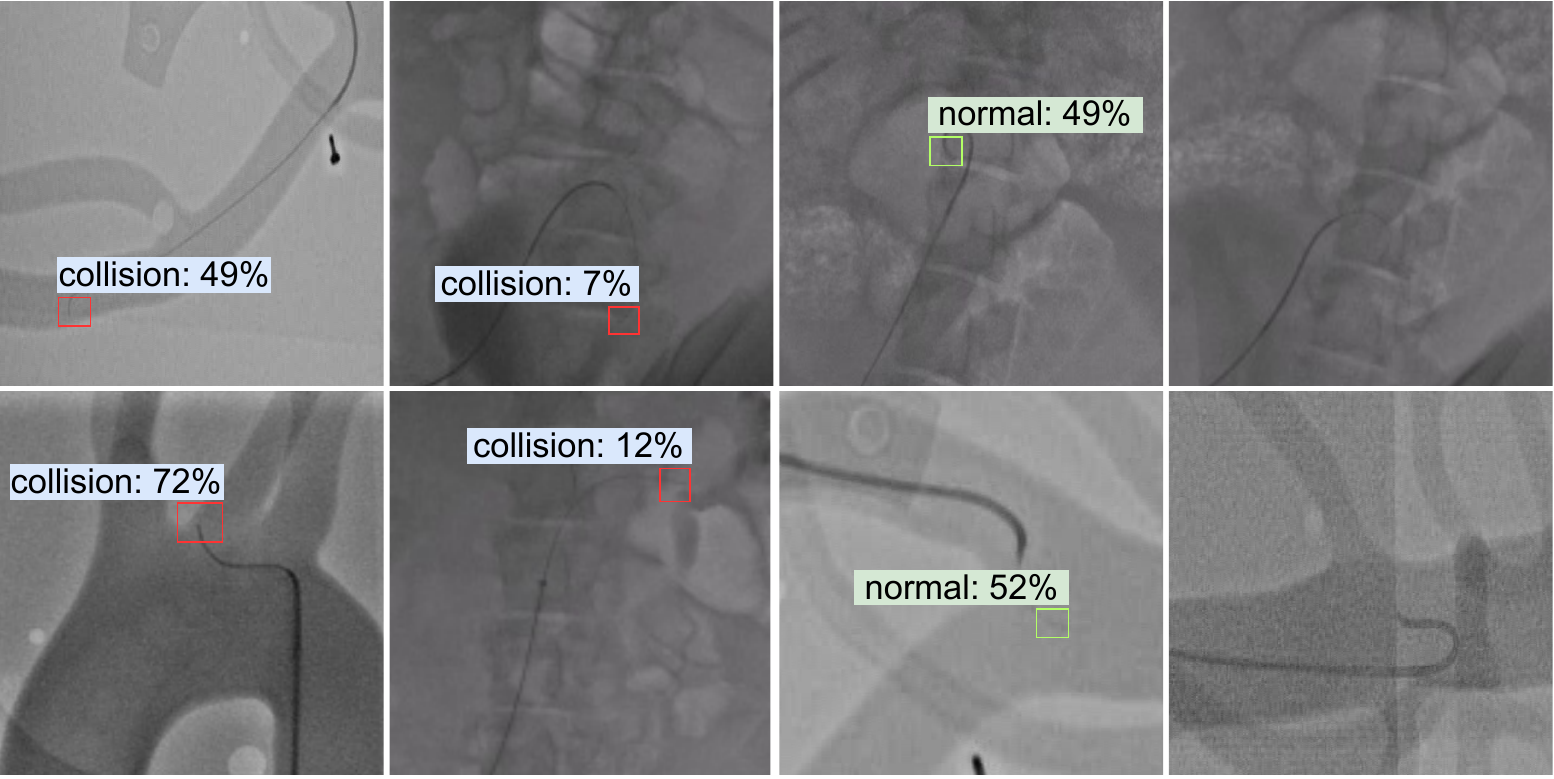}
    \caption{Qualitative results for the collision detection task. The first two columns visualize the collision results, the third column visualizes no collision cases, and the last column visualizes a failure case where the tip was not detected.}
    \label{fig:task_collision_vls}
\end{figure}

\textbf{Results.} 
Table~\ref{table:collision_results} shows the collision detection results. This table indicates that tiny object detectors such as Yolov~\cite{shi2023yolov} and EFF~\cite{gong2021effective} achieve higher accuracy compared to other normal object detectors. Furthermore, we observe that the performance of all methods remains relatively low. This highlights the challenges that lie ahead for collision detection in endovascular intervention. Fig.~\ref{fig:task_collision_vls} shows detection examples where EFF~\cite{gong2021effective} has difficulty when detecting collision between the catheter and the blood vessel. 

\textbf{Discussion.} Compared to traditional object detection results on vision datasets such as COCO~\cite{fang2023eva,dai2021dynamic,wang2023yolov7,li2022mvitv2}, the collision detection results on our dataset are significantly lower, with the top mean AP being only $16.91$. The challenges of this task come from two factors. First, the tip of the catheter or guidewire is relatively small in X-ray images. Second, the imbalance between the \texttt{collision} and \texttt{normal} class makes the problem more difficult. Therefore, there is a need to develop special methods to address these difficulties. Future works may rely on attention mechanisms, transformer, or foundation model~\cite{he2023camouflaged,cao2023multimodal,li2023discriminative} to develop more sufficient endovascular collision detectors.

\subsection{Domain Adaptation}
Our dataset is sourced from two distinct environments: vascular \textit{phantom data} and \textit{animal data}. To assess the capacity for learning from phantom data and applying it to real data, we benchmark endovascular interventions under domain adaptation setups~\cite{farahani2021brief,csurka2017domain}. For each task, we train the model on the phantom data and then test them on real animal data. In practice, animal data is similar to human data we capture from humans, and it is much more challenging to perform tasks on real animal or human data.

\begin{table}[h]
\centering
\renewcommand
\tabcolsep{4.5pt}
\hspace{1ex}
\vskip 0.1 in
\resizebox{\linewidth}{!}{
\begin{tabular}{@{}lcccc@{}}
\toprule
Baseline & Venues & Accuracy & Precision &Recall\cr 
\midrule
RU-LSTM~\cite{furnari2019would}&CVPR 2019& 22.93  & 23.91  & 22.57 \\
TempAggRe~\cite{sener2020temporal}&ECCV 2020 & 17.16 & 18.41  &18.23  \\ 
Trans-SVNet~\cite{jin2022trans} & IJCARS 2022 & 19.06& 17.67  & 19.58 \\
AFFT~\cite{zhong2023anticipative} &WACV 2023& \textbf{25.67}  & \textbf{26.29}  & \textbf{26.33} \\
\bottomrule
\end{tabular}
}
\caption{\label{tab:adaptation} Catheterization anticipation results under domain adaptation setup. All methods are trained on phantom data and tested on animal data.}
\end{table}

\textbf{Anticipation Adaptation.} We use the same methods RU-LSTM~\cite{furnari2019would}, TempAggRe~\cite{sener2020temporal}, Trans-SVNet~\cite{jin2022trans}, and AFFT~\cite{zhong2023anticipative}) for anticipation adaptation experiments. Table~\ref{tab:adaptation} shows the results. Compared with the setup in Table~\ref{tab:act_ant}, we can see that there is a significant accuracy drop. This highlights the challenges of applying baseline methods in practical real-world scenarios, particularly when dealing with unforeseen situations in catheterization procedures.
\begin{table}[t]
\centering
\renewcommand
\tabcolsep{4.5pt}
\hspace{1ex}
\vskip 0.1 in
\resizebox{\linewidth}{!}{
\begin{tabular}{@{}lcccc@{}}
\toprule
Baseline & Venues & Accuracy & Precision &Recall\cr 
\midrule
TDN-ResNet50~\cite{wang2021tdn}  &CVPR 2021& 24.19&23.17&24.56    \\
TDN-ResNet101~\cite{wang2021tdn} &CVPR 2021& 25.62&24.52&25.68    \\
Video Swin Transformer~\cite{xia2022vision} &CVPR 2022& 28.79 &27.98&28.12    \\
BEVT ~\cite{wang2022bevt} &CVPR 2022& \textbf{31.22}&\textbf{30.48}&\textbf{31.79} \\
\bottomrule
\end{tabular}
}
\caption{\label{tab:Recog_adapt} Catheterization recognition results under domain adapatation setup.} 
\end{table}

\textbf{Recognition Adaptation.} We repeat the catheterization recognition task under the domain adaptation setup. 
Table~\ref{tab:Recog_adapt} shows the results when all baselines are trained on phantom data and tested on animal data. This table also demonstrates that training under domain adaption setup is very challenging, as compared to Table~\ref{tab:Recog} under normal setting, the accuracy drops approximately $30\%$.

\begin{table}[h]
\centering
\renewcommand
\tabcolsep{4.5pt}
\hspace{1ex}
\vskip 0.1 in
\resizebox{\linewidth}{!}{
\begin{tabular}{@{}lcccccc@{}}
\toprule
  & \multicolumn{3}{c}{\textbf{AP}} & \multicolumn{3}{c}{\textbf{mAP}} \\
\cmidrule(lr){2-4}\cmidrule(lr){5-7}
Baseline &  
Collision & Normal  & Mean & Collision & Normal & Mean\cr 
\midrule
STEP~\cite{yang2019step}& 1.53 & 2.12 & 1.87  & 1.09 & 1.98 & 1.62  \\
YOWO~\cite{kopuklu2019you}& 2.12 & 4.11 & 3.09  & 1.97 & 3.68 & 2.92  \\
YOWO-Plus~\cite{yang2022yowo}& 1.18 & 1.43 & 1.21  & 1.07 & 1.26 & 1.09  \\
HIT~\cite{faure2023holistic}& 1.31 & 1.19 & 1.24  & 1.06 & 1.18 & 1.11  \\
\hline
Yolov$^*$~\cite{shi2023yolov} &7.31&8.92&8.09&6.28&7.49&7.21 \\
EFF$^*$~\cite{gong2021effective} &\textbf{8.27}&\textbf{9.16}&\textbf{8.19}&\textbf{7.61}&\textbf{8.29}&\textbf{7.88} \\
\bottomrule
\end{tabular}}
\vspace{-1ex}
\caption{\label{table:collision_adapt_results} Collision detection results under domain adaptation setup. All methods are trained on phantom data and tested on animal data. The symbol ($^*$) denotes tiny object detectors}
\end{table}

\textbf{Collision Detection Adaptation.} Table~\ref{table:collision_adapt_results} shows the results collision detection results under domain adaptation. We can see that under domain adaptation setup, most object detection methods achieve very low accuracy. Therefore, there is an immediate need to improve or design new methods that can detect the collision in real-time for endovascular catheterization procedures.

\begin{table}[h]
\centering
\renewcommand
\tabcolsep{4.5pt}
\hspace{1ex}
\vskip 0.1 in
\resizebox{\linewidth}{!}{
\begin{tabular}{@{}lcccccc@{}}
\toprule
Baseline & Dice Score & Jaccard Index & mIoU & Accuracy\cr 
\midrule
UNet~\cite{ronneberger2015u}       & 26.58    & 31.38     & 12.13    & 46.07\\ 
TransUNet~\cite{chen2021transunet} & 16.16    & 24.19     & 17.23    & 33.61 \\ 
SwinUNet~\cite{ambrosini2017fully} & 17.41    & \textbf{38.14}     & 7.52    & 40.79\\
SSL~\cite{kongtongvattana2023shape} & 26.91    & 32.04     & \textbf{18.72}    & 42.44\\
SegViT~\cite{zhang2022segvit}      & \textbf{30.74}    & {32.22}     & {11.46}    & \textbf{50.00} \\
\bottomrule
\end{tabular}}
\vspace{-1ex}
\caption{\label{table:seg_results_adapatation} Domain adaptation segmentation results.}
\vspace{-3ex}
\end{table}

\textbf{Segmentation Adaptation.} Table~\ref{table:seg_results_adapatation} shows the catheter and guidewire segmentation results when the networks are trained on phantom data and tested on animal data. Similar to other tasks under the domain adaptation setting, we observe a significant accuracy drop in all methods. Overall, SegViT~\cite{zhang2022segvit} still outperforms other segmentation methods. This shows that the vision transformer backbone may be potentially a good solution for this task. 

\section{Discussion}
\label{sec_conclusion}
We introduce CathAction, a large-scale dataset for endovascular intervention tasks, encompassing annotated groundtruth for segmentation, action understanding, and collision detection. While CathAction marks a significant advancement in endovascular interventions, it is important to acknowledge certain limitations. First, despite its comprehensiveness, the dataset may not encompass every possible clinical scenario and could potentially lack representation for rare or outlier cases~\cite{murtaza2023synthetic}. Second, our work currently benchmarks vision-based methods, which exhibit insufficient accuracy, and persisting challenges in generalizability and adaptability to real-world scenarios are evident. This is highlighted by the results presented in Section~\ref{sec_task} for all catheterization anticipation, recognition, segmentation, and collision detection tasks. Thirdly, we mostly utilize metrics from the vision community to evaluate the results. These metrics may not fully reflect the clinical needs, and the continuous refinement of evaluation metrics and exploration of potential interdependencies among tasks would demand further research~\cite{xu2023exploring}. 

From our intensive experiments, we see several research directions that benefit from our large-scale datasets: \textit{i)} There is an immediate need to develop more advanced methods for catheterization anticipation, recognition, collision detection, and action understanding, especially under domain adaption setup. Future work can explore the potential of graph neural networks~\cite{li2021adaptive,wu2021anticipating}, temporal information~\cite{furnari2019would,zhong2023anticipative}, multimodal or transfer learning~\cite{tufek2019human} to improve the accuracy and reliability of the methods. \textit{ii)} Currently, we address endovascular intervention tasks independently, future work can combine those tasks and tackle them simultaneously (e.g., the anticipation and collision detection tasks can be jointly trained). This would make the research outputs more useful in clinical practice~\cite{abdelaziz2019toward}. Finally \textit{iii)}, given the fact that CathAction is a large-scale dataset, it can be used to train a foundation model for endovascular interventions or related medical tasks~\cite{nguyen2023lvm}.

\section{Conclusion}
We introduce CathAction as a large-scale dataset for endovascular intervention research, offering the largest and most comprehensive benchmark to date. With intensive annotated data, CathAction addresses crucial limitations in existing datasets and helps connect computer vision with healthcare tasks. By providing a standardized dataset with public code and public metrics, CathAction promotes transparency, reproducibility, and the collective exploration of different tasks in the field. Our code and dataset are publicly available to encourage further study.

\bibliographystyle{class/IEEEtran}
\bibliography{class/egbib}
   
\end{document}